# OPTIMAL FLEXURAL DESIGN OF FRP-REINFORCED CONCRETE BEAMS USING A PARTICLE SWARM OPTIMIZER

M. S. INNOCENTE[1], LL. TORRES[1], X. CAHÍS[1], G. BARBETA[1], and A. CATALÁN[2]

[1] Department of Mechanical and Construction Engineering, University of Girona, Girona, Spain
[2] Department of Construction and Manufacturing Engineering, University of Oviedo, Gijón, Spain

**Keywords**: fibre-reinforced polymers, concrete, rectangular beams, ACI 440.1 R-06, least-cost section design, particle swarm optimization, constraint-handling techniques.

## 1 INTRODUCTION

The problems that arise when steel-reinforced concrete members are exposed to aggressive environments such as the corrosion of the reinforcement have been widely discussed in the literature. Different solutions have been proposed, being the use of fibre-reinforced polymers (FRPs) in replacement of steel one of the newest. However, the precise features of FRPs are still far from being standardized, so that their properties greatly depend on the manufacturer.

Although the flexural design of FRP-reinforced members resembles that of the steel-reinforced ones, the mechanical properties of these materials differ. Thus, FRPs exhibit high tensile strength only in the direction of the fibres, do not exhibit yielding, and present bond stress-bond slip behaviour different from that of steel. In the absence of a code, the design of FRP-reinforced concrete beams within this paper is based on the recommendations in the recently issued ACI 440.1 R-06: "Guide for the Design and Construction of Structural Concrete Reinforced with FRP bars" [1]. Since the use of FRP-reinforced concrete is in its early stages, both analytical and experimental results are still scarce. Besides, the properties of the FRP bars used along the different experiments typically differ from one another due to the lack of standardization. In agreement with the above-mentioned recommendations [1], the flexural design is limited here to rectangular beams with tensile reinforcement only.

The design of the cross-section of an FRP-reinforced concrete beam is an iterative process of estimating both its dimensions and the reinforcement ratio, followed by the check of the compliance of a number of strength and serviceability constraints. The process continues until a suitable solution is found. Since there are infinite solutions to the problem, it appears convenient to define some optimality criteria so as to measure the relative goodness of the different solutions. This paper intends to develop a preliminary least-cost section design model that follows the recommendations in the ACI 440.1 R-06 [1], and uses a relatively new artificial intelligence (AI) technique called "particle swarm optimization" (PSO) to handle the optimization tasks. The latter is based on the intelligence that emerges from the low-level interactions among a number of relatively non-intelligent individuals within a population.

The optimal design of steel-reinforced concrete structures using classical and modern optimization methods has already received attention from numerous researchers (e.g. [2, 3, 4, 5]). In contrast, to our knowledge, the optimal design of FRP-reinforced members and the use of PSO algorithms to deal with the optimal design of concrete structures have not been addressed in the literature.

## 2 FRP-REINFORCED CONCRETE FLEXURAL SECTION DESIGN ACCORDING TO THE ACI 440.1 R-06 RECOMMENDATION

Although a detailed discussion on the flexural design procedure recommended in the ACI 440.1 R-06 [1] is beyond the scope of this paper, a brief enumeration of the main steps is presented hereafter in order to depict the procedure.

### 2.1 Step 1: Estimate the appropriate cross-sectional dimensions of the beam

Traditionally, the cross-sectional dimensions of a concrete beam reinforced with steel are initially estimated according to some standard ratio between the depth of the beam and the span between its supports. For a first estimation of the cross-sectional dimensions of an FRP-reinforced concrete beam, higher values are recommended because of the comparatively lower stiffness of the FRP bars.

## 2.2 Step 2: Compute the factored uniform load and ultimate moment

The uniformly distributed dead load ($w_{DL}$) can be computed as the superimposed dead load ($w_{SDL}$) plus the self-weight ($w_{SW}$) of the beam:

$$w_{DL} = w_{SDL} + w_{SW} \tag{1}$$

The "ultimate moment" ($M_u$) is calculated according to the boundary conditions of the beam using the "factored uniform load", which is computed as follows:

$$w_u = 1.2 \cdot w_{DL} + 1.6 \cdot w_{LL} \tag{2}$$

## 2.3 Step 3: Compute the design tensile strength and rupture strain of the FRP bars

The guaranteed tensile strength ($f_{fu}^*$) and tensile rupture strain ($\varepsilon_{fu}^*$) provided by the manufacturer should be considered as initial properties, which do not include the effects of long-term exposure to the environment. Thus, the design tensile strength ($f_{fu}$) and the design rupture strain ($\varepsilon_{fu}$) are computed as follows:

$$f_{fu} = C_E \cdot f_{fu}^* \quad ; \quad \varepsilon_{fu} = C_E \cdot \varepsilon_{fu}^* \tag{3}$$

The coefficient $C_E$ is an environmental reduction factor, which can be taken from Table 7.1 in [1].

## 2.4 Step 4: Compute the balanced reinforcement ratio

The balanced reinforcement ratio ($\rho_{fb}$) is the threshold between the crushing failure modes: the failure occurs via the concrete crushing if the reinforcement ratio is greater than the balanced ratio, while the failure occurs via the FRP bars if the reinforcement ratio is under the threshold.

$$\rho_{fb} = 0.85 \cdot \beta_1 \cdot \frac{f_c^{'}}{f_{fu}} \cdot \frac{E_f \cdot \varepsilon_{cu}}{E_f \cdot \varepsilon_{cu} + f_{fu}} \tag{4}$$

Where:
- $\beta_1$ : factor taken as 0.85 for concrete strength up to 27.58 MPa; for strength above this value, this factor is reduced continuously at a rate of 0.05 per each 6.895 MPa of strength in excess of 27.58 MPa, but it is not taken less than 0.65
- $f_c^{'}$ : specified compressive strength of concrete [MPa]
- $E_f$ : guaranteed modulus of elasticity of FRP [MPa]
- $\varepsilon_{cu}$ : ultimate strain in concrete

## 2.5 Step 5: Adopt the reinforcement ratio

Because the FRP materials do not exhibit yielding when loaded in tension, the behaviour is characterized by a linearly elastic stress-stain relationship until failure, which results in the failure of the member without warning. Thus, as opposed to steel-reinforced concrete members, it is usually preferred to over-reinforce FRP-reinforced concrete members, so that the concrete crushing controls the failure. Nevertheless, the ACI 440.1 R-06 recommendation considers the design procedure for both concrete members under and over-reinforced with FRP bars. A possible approach to a traditional design which somehow considers the final cost of the beam could be to adopt the reinforcement ratio equal to the minimum as a first estimation, and then start checking the strength and serviceability requirements. Then, the ratio is iteratively increased when necessary.

$$\rho_f = \frac{A_f}{b \cdot d} \tag{5}$$

Where $A_f$ stands for the area of the FRP-reinforcement, and *d* for the distance from the fibre subject to the greatest tensile stress to the centre of gravity of the reinforcement.

Note that the tensile strength of the bars decreases as the diameter increases. Hence it is necessary not only to define the reinforcement ratio but also to decide upon the numbers and diameters of the bars, with the corresponding update of the reinforcement ratio. The minimum ratio is as follows:

$$\rho_{f,\min} = \max\left(0.4070 \cdot \frac{\sqrt{f'_c\ [MPa]}}{f_{fu}\ [MPa]}\ ,\ \frac{2.256}{f_{fu}\ [MPa]}\right) \tag{6}$$

## 2.6 Step 6: Check the required flexural strength

The check of the flexural strength depends on whether the section is under or over-reinforced.

× If $\rho_f \leq \rho_{fb}$

$$M_n\ [KN \cdot m] = 1000 \cdot A_f\ [m^2] \cdot f_{fu}\ [MPa] \cdot d\ [m] \cdot \left[1 - \frac{\beta_1}{2} \cdot \frac{\varepsilon_{cu}}{\varepsilon_{cu} + \varepsilon_{fu}}\right] \tag{7}$$

$$\phi = 0.55 \tag{8}$$

× If $\rho_f > \rho_{fb}$

$$f_f = \sqrt{\frac{(E_f \cdot \varepsilon_{cu})^2}{4} + \frac{0.85 \cdot \beta_1 \cdot f'_c \cdot E_f \cdot \varepsilon_{cu}}{\rho_f}} - 0.5 \cdot E_f \cdot \varepsilon_{cu} \tag{9}$$

$$M_n\ [KN \cdot m] = 1000 \cdot A_f\ [m^2] \cdot f_f\ [MPa] \cdot \left[d\ [m] - \frac{A_f \cdot f_f}{1.7 \cdot f'_c \cdot b}\right] \tag{10}$$

$$\phi = \begin{cases} 0.65 & \text{if } \rho_f > 1.4 \cdot \rho_{fb} \\ 0.3 + 0.25 \cdot \dfrac{\rho_f}{\rho_{fb}} & \text{otherwise} \end{cases} \tag{11}$$

Finally, the flexural strength requirement is met if:

$$\phi \cdot M_n \geq M_u \tag{12}$$

If this condition is not met, the reinforcement ratio needs to be increased.

## 2.7 Step 7: Check the crack width

Since FRP-reinforced concrete members present relatively small stiffness after cracking, permissible deflections under service loads may control the design. Given that the FRP bars are corrosion resistant, aesthetics might be the primary reason for the crack-width limitation. The ACI 440.1 R-06 [1] recommends using the limitations of the crack width allowed by the Canadian Standards Association: 0.5 mm for exterior exposure, and 0.7 mm for interior exposure.

2.7.1 Compute the stress in the FRP reinforcement

$$n_f = \frac{E_f}{E_c}\ ;\quad k = \sqrt{(\rho_f \cdot n_f)^2 + 2 \cdot \rho_f \cdot n_f} - \rho_f \cdot n_f \tag{13}$$

$$f_f\ [MPa] = \frac{M_{DL+LL}\ [KN \cdot m]}{1000 \cdot A_f\ [m^2] \cdot d\ [m] \cdot (1 - k/3)} \tag{14}$$

Where $M_{DL+LL}$ is the moment acting on the section due to the dead and live loads.

2.7.2 Compute the crack width

Calculate the distance $d_c = h - d$ from the extreme tension fibre to the centreline of the reinforcement and compute bar spacing $s$ according to (15), and estimate the crack width as shown in (16).

$$s = \frac{b - 2 \cdot d_c}{\text{number of bars} - 1} \tag{15}$$

$$w = \frac{2}{E_f} \cdot \frac{h - k \cdot d}{d \cdot (1 - k)} \cdot k_b \cdot f_f \cdot \sqrt{d_c^2 + \left(\frac{s}{2}\right)^2} \tag{16}$$

Where $k_b$ is a coefficient that accounts for the degree of bond between the FRP bars and the surrounding concrete. If this value is unknown, the ACI 440.1 R-06 [1] suggests using $k_b = 1.4$.

### 2.8  Step 8: Check the permissible short and long-term deflections

First of all, the gross moment of inertia of the section is computed:

$$I_g = \frac{b \cdot h^3}{12} \tag{17}$$

When the applied moment exceeds the cracking moment, cracking occurs causing a reduction in the stiffness. The moment of inertia is computed on the cracked section:

$$I_{cr} = \frac{b \cdot d^3}{3} \cdot k^3 + n_f \cdot A_f \cdot d^2 \cdot (1-k)^2 \tag{18}$$

The effective moment of inertia $I_{cr} \leq I_e \leq I_g$ depends on the magnitude of the applied moment. The use of the following equation for the computation of the effective moment of inertia—which also takes into account the tension-stiffening effect—is recommended in the ACI 440.1 R-06 [1]:

$$(I_e)_{DL+LL} = \left(\frac{M_{cr}}{M_{DL+LL}}\right)^3 \cdot 0.2 \cdot \frac{\rho_f}{\rho_{fb}} \cdot I_g + \left[1 - \left(\frac{M_{cr}}{M_{DL+LL}}\right)^3\right] \cdot I_{cr} \leq I_g \tag{19}$$

Where the cracking moment ($M_{cr}$) is given by:

$$M_{cr}\,[\text{KN}\cdot\text{m}] = \frac{1240 \cdot \sqrt{f'_c} \cdot I_g}{h} \tag{20}$$

The immediate deflection due to dead and live loads $(\Delta i)_{DL+LL}$ is computed using the effective moment of inertia, while the deflection due to dead and live loads independently from one another are computed as a weighted average:

$$(\Delta i)_{DL} = \frac{w_{DL}}{w_{DL} + w_{LL}} \cdot (\Delta i)_{DL+LL} \tag{21}$$

$$(\Delta i)_{LL} = \frac{w_{LL}}{w_{DL} + w_{LL}} \cdot (\Delta i)_{DL+LL} \tag{22}$$

Finally, the long term deflection is computed as follows:

$$\Delta_{LT} = (\Delta i)_{LL} + 0.6 \cdot \xi \cdot \left[(\Delta i)_{DL} + (\%) \cdot (\Delta i)_{LL}\right] \tag{23}$$

Where (%) represents the percentage of the live load that can be considered sustained (typically set to 20%), and $\xi = 2$ for more than 5 years.

### 2.9  Step 9: Check the creep rupture stress limits

Compute the moment due to all sustained loads as follows:

$$M_s = \frac{w_{DL} + (\%) \cdot w_{LL}}{w_{DL} + w_{LL}} \cdot M_{DL+LL} \tag{24}$$

Then, compute the sustained stress level in the FRP bars:

$$f_{f,s}\,[\text{MPa}] = \frac{M_s\,[\text{KN}\cdot\text{m}]}{1000 \cdot A_f\,[\text{m}^2] \cdot d\,[\text{m}] \cdot (1-k/3)} \tag{25}$$

Finally, the creep rupture stress limit requirement is met if

$$f_{f,s} \leq F_{f,s} \tag{26}$$

Where $F_{f,s}$ is the creep rupture stress limit, which can be taken from Table 8.2 in [1].

## 3 OPTIMAL DESIGN

The guidelines for the design of concrete structures that can be found in the codes and recommendations are only concerned with the satisfaction of some strength and serviceability constraints. Among the infinite designs which meet those requirements, it is worthwhile seeking the ones that result in the minimal cost of the structure. As it can be observed from the previous section, the standard procedure for the design is quite tedious, allowing the check of a few configurations only. Of course, it is always possible to implement the standard procedure in a computer program, so that numerous designs can be examined, choosing the one associated to the minimal cost. However, every run of the program would be independent from the others, thus neglecting useful information gained from previous experiences. The present paper aims to develop a model for the least-cost flexural design of FRP-reinforced concrete beams, subject to strength and serviceability constraints.

The formulation of the problem must express precisely what is desired to be solved, for which the definition of a function of the design variables that successfully measures the idea of optimality is critical. In addition, setting the constraints in agreement with the real problem is also critical, since otherwise the solution might be infeasible. Thus, the problem needs to be reduced to a function of the design variables which has to be optimized, plus a number of constraints that limit the feasible regions of the search-space. The function to be optimized in a mathematical model is traditionally called "cost function", although other names such as "objective function" and "fitness function" are frequent in the literature. Due to the nature of the metaphor which the PSO method was originally inspired on (refer to the next section), the function to be minimized is referred to as the "conflict function". In summary, a well posed optimization problem requires the definition of the conflict function, of the search-space, and of the constraints that delimitate the feasible part of the search-space.

Let $\mathcal{S}$ be the search-space, and $\mathcal{F} \subseteq \mathcal{S}$ its feasible part. A minimization problem consists of finding the vector $\hat{\mathbf{x}} \in \mathcal{F}$ such that $f(\hat{\mathbf{x}}) \leq f(\mathbf{x})$, where $\mathbf{x} \in \mathcal{F}$. Therefore, the formulation can be as follows:

$$\text{Minimize } f(\mathbf{x}) \quad \text{subject to } \mathbf{x} \in \mathcal{F} \tag{27}$$

More precisely, the problem can be rewritten as:

$$\text{Minimize } f(\mathbf{x}) \quad \text{subject to } \begin{cases} g_j(\mathbf{x}) \geq 0 & ; \quad j = 1, \ldots, q \\ g_j(\mathbf{x}) = 0 & ; \quad j = q+1, \ldots, m \end{cases} \tag{28}$$

Where: - $\mathcal{S}$ is the search-space
- $\mathbf{x} \in \mathcal{S}$ is the vector of object variables
- $f(\cdot): \mathcal{S} \to \mathcal{E} \subseteq \mathcal{R}$ is the function to be optimized, where $\mathcal{R}$ is the set of real numbers
- $g_j(\cdot)$ are the constraint functions

Therefore, given a design of the section of an FRP-reinforced concrete beam, a function to compute its cost needs to be developed, and the problem consists of minimizing such a function. The strength and serviceability requirements are formulated as constraints to the minimization problem.

While a detailed discussion about the differences between "traditional methods" and the so-called "modern heuristics" is beyond the scope of this paper, it is noted here that modern heuristics such as evolutionary algorithms (EAs) and particle swarm optimizers (PSOs) are AI-based techniques which perform a parallel search without using gradient information, resulting in robust and general-purpose rather than problem-specific algorithms that require few or no adaptation to handle different problems.

## 4 PARTICLE SWARM OPTIMIZATION

### 4.1 Introduction

The PSO paradigm was originally developed by social-psychologist James Kennedy and electrical-engineer Russell Eberhart in 1995 [6]. Despite the fact that the paradigm is mainly used in practice to handle optimization problems, some principles underlying simulations of socio-cognitive phenomena were of great influence for its development. In fact, although the method was inspired by previous bird-flock simulations, such simulations were framed within the field of social psychology, under the socio-cognitive view of mind (i.e. thinking and intelligence as social phenomena).

Experiments carried out in social psychology such us those of Sheriff, Asch and Bandura (all quoted in [7]) suggest that whenever people interact, they become more similar to one another. This is because they tend to seek agreement by imitating the most successful ones. This is the key concept underlying the PSO paradigm. Some of the studies concerning social phenomena were undertaken by simulating the behaviour observed in some social animals such us bird-flocks, fish-herds and social insects (e.g. ants, termites, and bees). A well-known simulation of bird-flocks was developed by Reynolds (quoted in [7]), who proposed three basic rules for each bird to follow: (1) pull away before crashing into another bird; (2) try to match the neighbours' velocities; (3) try to move towards the centre of the flock.

Another influential work was that of Heppner and Grenander (quoted in [6, 7]), who observed the critical issue that natural bird-flocks do not have a leader. In other words, there is no central control! Heppner and Grenander implemented a simulation similar to that of Reynolds, but now the birds were also attracted to a roost, and an occasional, random force was implemented seldom deflecting the birds' direction, thus resembling a gust of wind. The fact that Heppner and Grenander's artificial birds (quoted in [6, 7]) were attracted to a roost (or to a food source) led Kennedy et al. [6] to think of optimization. However, those simulations profited from knowing the location of the "roost" in advance, while both real birds and the PSO algorithm seek "food" without any prior knowledge about its location. Instead, they carry out a parallel exploration, profiting from sharing the information acquired by every individual. Thus, the PSO paradigm was originated on the simulation of a simplified social milieu, where individuals were thought of as collision-proof birds flying over an *n*-dimensional search-space. Refer to Kennedy et al. [6] for further details.

Kennedy et al. [7] suggest that the behaviour of the individuals within a population can be summarized in terms of three principles:

1. Evaluate: An organism evaluates the environment by evaluating the stimuli perceived by its sensors in order to decide the proper reaction.

2. Compare: Once the perceived stimuli are evaluated, it is not straightforward to tell good from bad. Experiments in social psychology suggest that individuals judge themselves by comparing to others. In other words, by telling better from worse rather than good from bad.

3. Imitate: An individual only imitates those whose performances are superior or somehow desirable.

These three processes are implemented within the PSO paradigm with remarkable success.

Within the broad field of AI, a relatively new branch called artificial life (AL) encompasses all the paradigms inspired on phenomena observed in biological organisms. Some AL paradigms are viewed as a route to AI, where intelligent behaviour is not programmed but emerges from lower level, local interactions. EAs and PSOs, which are examples of such paradigms, are population-based methods that lean on a population of individuals spread over the design space, where each individual returns a candidate solution. Both methods are also similar in that they are not specifically programmed to perform optimization tasks. Instead, their ability to optimize is an emergent property, which is the most distinctive feature of the AL paradigms. "Emergent" property means that it is formed from lower level individual-to-individual interactions that appear to be unrelated to the higher level resulting property.

**4.2  BASIC PARTICLE SWARM OPTIMIZER**

The basic equations that rule the trajectories of the particles are as follows:

$$v_{ij}^{(t)} = w^{(t)} \cdot v_{ij}^{(t-1)} + iw^{(t)} \cdot U_{(0,1)} \cdot \left(pbest_{ij}^{(t-1)} - x_{ij}^{(t-1)}\right) + sw^{(t)} \cdot U_{(0,1)} \cdot \left(gbest_{j}^{(t-1)} - x_{ij}^{(t-1)}\right) \quad (29)$$

$$x_{ij}^{(t)} = x_{ij}^{(t-1)} + v_{ij}^{(t)} \quad (30)$$

Where:
- $x_{ij}^{(t)}$ : coordinate *j* of the position of particle *i* at time-step *t*
- $v_{ij}^{(t)} = \Delta x_{ij}^{(t)}$ : component *j* of the velocity of particle *i* at time-step *t*
- $w^{(t)}, iw^{(t)}, sw^{(t)}$ : inertia, individuality, and sociality weights at time-step *t*
- $U_{(0,1)}$ : random number generated from a uniform distribution in the range $[0,1]$, re-sampled anew each time it is referenced
- $pbest_{ij}^{(t-1)}$, $gbest_{j}^{(t-1)}$ : coordinate *j* of the best position found by particle *i* and by the whole swarm, respectively, up to time-step $(t-1)$

As can be seen from equation (29), a particle's velocity at every time-step is computed as the velocity at the previous one (weighted by the inertia weight), altered by two components: one related to the particle's memory of its best previous experience, and the other related to the swarm's memory of its best previous experience. In turn, the random weights introduce creativity into the system. Since the latter are re-sampled anew for each time-step, for each particle, for each component, and for each term of equation (29), the particles display uneven, zigzagging trajectories. This results in better exploration. Besides, re-sampling the random weights anew for the individuality and the sociality terms makes each particle alternate between a more "self-confident behaviour" (favouring exploration) and a more conformist one (favouring exploitation) without any of them taking the lead for too long.

The original algorithm, which considered $w^{(t)} = 1$, presented a serious problem: the particles tended to diverge rather than cluster, so that a so-called "explosion" took place. It was found that clumping the components of the particles' velocities effectively controlled the explosion, and the particles ended up clustering around a solution. While the dynamics of the system and the reasons for the explosion are beyond the scope of this paper, it is merely mentioned here that they were found to be related to both the relative importance given to the second and third terms over the first one in equation (29), and to the random weights. Some important theoretical works in this regard were performed by Kennedy et al. [7], Clerc et al. [8], and Engelbrecht [9]. Although the incorporation of the inertia weight as proposed by Shi et al. [10] (or of the constriction factor as proposed by Clerc et al. [8]) are able to control the explosion—and to favour exploitation—, clamping the components of the particles' velocities still appear to be desirable.

There are two main versions of the algorithm: the local PSO and the global PSO. In the former, the trajectory of a particle at a given time-step can be influenced only by its own experiences and by those of a few other particles comprising its neighbourhood. Since the neighbourhoods are defined so that they overlap, the experiences can be spread over the whole population. Instead, the global PSO considers a single neighbourhood, so that every particle is connected to all the others. The information is spread faster in the second case. This paper is only concerned with the global PSO.

### 4.3   CONSTRAINT-HANDLING TECHNIQUES

Although different techniques have been proposed in the literature to deal with the constraints, the appropriate choice appears to be problem-dependent. Two constraint-handling techniques are implemented within this work: the "preserving feasibility" technique and the "penalization" method. An extensive discussion on the different techniques and their variations is beyond the scope of this paper. Refer to Innocente [11], Engelbrecht [9], and Hu et al. [12] for further details.

While the development of a general-purpose PSO optimizer can be found in Innocente [11], refer to Innocente et al. [13] for a fairly extensive overview of the paradigm.

## 5   FRP-REINFORCED CONCRETE FLEXURAL OPTIMAL DESIGN USING A PSO

Two different optimizers are used in the experiments carried out along this paper:

1. GP-PSO$^{(s.d.w.)}$, as described in [11] (page 404), with the incorporation of the "preserving feasibility" technique to handle the constraints. The (real-valued) design variables are the width and height of the beam, and the reinforcement is computed in a deterministic fashion following the ACI 440.1 R-06 [1].
2. A slight modification of the former (the "sigmoidly" time-decreasing inertia weight is replaced by a constant one equal to 0.8), and the constraint-handling technique now consists of the "penalization" method. The penalization coefficient is linearly time-increasing from $k^{(t=1)} = 10^5$ to $k^{(t=t_{max})} = 10^{10}$. The design variables are the width and height of the beam, as well as the number and diameter of the bars. Thus, the search-space is now 4-dimensional, where two dimensions are real-valued and the other two are discrete. The strength and serviceability requirements, as well as the geometrical limitations, are formulated as constraints to the optimization problem.

A maximum number of time-steps equal to 10000 and a swarm-size equal to 35 particles are set for both optimizers. As a first step towards a more sophisticated model, the design here is limited to section flexural design, and to uniform diameter selection. In agreement with the ACI 440.1 R-06 [1], no compression reinforcement, a single layer of tensile reinforcement, and rectangular cross-sections are considered.

## 6  DESIGN EXAMPLES

A few examples of the application of the methodology depicted in section 2 embedded into the two proposed optimizers are presented within this section. The aim is to show their suitability and efficacy to deal with this over-simplified optimal design problem. The algorithms are applied to the least-cost design of a pin jointed GFRP-reinforced concrete beam, loaded with a uniformly distributed load. Only the flexural design of the cross-section associated to the maximum moment is carried out at this initial stage. The cost of the member is computed as follows:

$$C = c_1 \cdot (b \cdot h) + c_2 \cdot (b + 2 \cdot h) + c_3 \cdot n \tag{31}$$

Where: - $b$, $h$, $n$ : width and height of the beam, respectively, and number of bars
- $c_1$ : cost per m$^3$ of concrete, which is set to 100
- $c_2$ : cost per m$^2$ of shuttering relative to the cost of the concrete
- $c_3$ : cost per m of bar (note that this depends on the bar selected in the design)

Three different set of costs have been considered, which already include the costs of the materials, transport and formwork. The required data considered in the examples are as follows:

$L = 5\,\text{m}$ ; $w_{SDL} = 8\,\dfrac{\text{KN}}{\text{m}}$ ; $w_{LL} = 7\,\dfrac{\text{KN}}{\text{m}}$ (20% sustained); $\gamma_c = 24\,\dfrac{\text{KN}}{\text{m}^3}$ (concrete density); $\Delta_{LT\max} = \dfrac{L}{240}$ ; $\xi = 2$ ; $f'_c = 30\,\text{MPa}$ ; $w_{\max} = 0.7\,\text{mm}$ ; $E_f = 44800\,\text{MPa}$ ; $E_c = 26016.8\,\text{MPa}$ ; $\varepsilon_{cu} = 0.003$ ; $CE = 0.8$ ; $b_{\min} = 20\,\text{cm}$ ; $b_{\max} = 100\,\text{cm}$ ; $h_{\min} = 20\,\text{cm}$ ; $h_{\max} = 200\,\text{cm}$ .

The minimum free gap between bars is set to $\max(1.4 \cdot \phi_b,\ 3\,\text{cm})$, while the minimum cover is set to $\max(2.5 \cdot \phi_b,\ 4\,\text{cm})$. The features of the GFRP-bars adopted within this work are as shown in **Table 1**, while the application of the two algorithms returns the designs showed in **Table 2**.

**Table 1**  FRP-bars' guaranteed tensile strength ($f_{fu}^*$) and cost of the materials

|  |  |  | CASE A | CASE B | CASE C |  |
|---|---|---|---|---|---|---|
|  | Concrete |  | 100 | 100 | 100 |  |
|  | Shuttering |  | 25 | 2.95 | 35 |  |
| Size | Diameter $\phi_b$ [mm] | $f_{fu}^*$ [MPa] |  |  |  |  |
| #2 | 6.35 | 825 | 0.498361392 | 0.58630752 | 0.274098766 | COSTS |
| #3 | 9.53 | 760 | 1.099572057 | 0.514049937 | 0.76970044 |  |
| #4 | 12.70 | 690 | 1.543445568 | 1.815818315 | 0.848895062 |  |
| #5 | 15.88 | 655 | 2.328630417 | 2.739565196 | 1.280746729 |  |
| #6 | 19.05 | 620 | 3.285252527 | 3.285252527 | 1.80688889 |  |
| #7 | 22.23 | 586 | 4.419347369 | 5.199232198 | 2.430641053 |  |
| #8 | 25.40 | 550 | 5.72378227 | 6.733861494 | 3.148080249 |  |
| #9 | 28.65 | 517 | 7.241397324 | 8.519290969 | 3.982768528 |  |

**Table 2**  Optimal designs obtained by the two proposed optimizers for each of the three cost sets, where the height is limited to a maximum value of 200 cm.

| CASE | $b$ [m] | $h$ [m] | $n \times \phi_b$ | COST | | | |
|---|---|---|---|---|---|---|---|
|  |  |  |  | Concrete | Shuttering | Reinforcement | TOTAL |
| **FIRST OPTIMIZER** | | | | | | | |
| A | 0.2124 | 0.5346 | 3 x #6 | 11.3547 | 32.0395 | 9.8558 | **53.2499** |
| B | 0.2124 | 0.5346 | 3 x #6 | 11.3547 | 3.7807 | 11.5950 | **26.7304** |
| C | 0.2401 | 0.4883 | 3 x #7 | 11.7227 | 42.5823 | 7.2919 | **61.5970** |
| **SECOND OPTIMIZER** | | | | | | | |
| A | 0.2124 | 0.5346 | 3 x #6 | 11.3543 | 32.0387 | 9.8558 | **53.2488** |
| B | 0.2124 | 0.5346 | 3 x #6 | 11.3546 | 3.7806 | 11.5950 | **26.7302** |
| C | 0.2401 | 0.4882 | 3 x #7 | 11.7217 | 42.5792 | 7.2919 | **61.5928** |

If for some technical reason the height were limited to 35 cm, then: $b_{\min} = 20\,\text{cm}$ ; $b_{\max} = 100\,\text{cm}$ ; $h_{\min} = 20\,\text{cm}$ ; $h_{\max} = 35\,\text{cm}$ . The results obtained for this case are shown in **Table 3**:

**Table 3** Optimal designs obtained by the two proposed optimizers for each of the three cost sets, where the height is limited to a maximum value of 35 cm.

| CASE | $b$ [m] | $h$ [m] | $n \times \phi_b$ | COST | | | |
|---|---|---|---|---|---|---|---|
| | | | | Concrete | Shuttering | Reinforcement | TOTAL |
| **FIRST OPTIMIZER** | | | | | | | |
| A | 0.5067 | 0.3482 | 9 x #6 | 17.6414 | 30.0756 | 29.5673 | **77.2843** |
| B | 0.8801 | 0.3447 | 21 x #3 | 30.3362 | 4.6300 | 10.7950 | **45.7612** |
| C | 0.5067 | 0.3482 | 9 x #6 | 17.6414 | 42.1059 | 16.2620 | **76.0093** |
| **SECOND OPTIMIZER** | | | | | | | |
| A | 0.5067 | 0.3500 | 7 x #7 | 17.7354 | 30.1682 | 30.9354 | **78.8390** |
| B | 0.5067 | 0.3500 | 9 x #6 | 17.7341 | 3.5597 | 29.5673 | **50.8611** |
| C | 0.4776 | 0.3500 | 6 x #8 | 16.7174 | 41.2174 | 18.8885 | **76.8233** |

The evolution of the particles' positions over the 2-dimensional space searched by the first optimizer is shown in **Fig. 1**. As can be observed one sub-swarm seeks the best solution (minimizer) and the other the worst solution (maximizer). The latter is only used for the termination conditions.

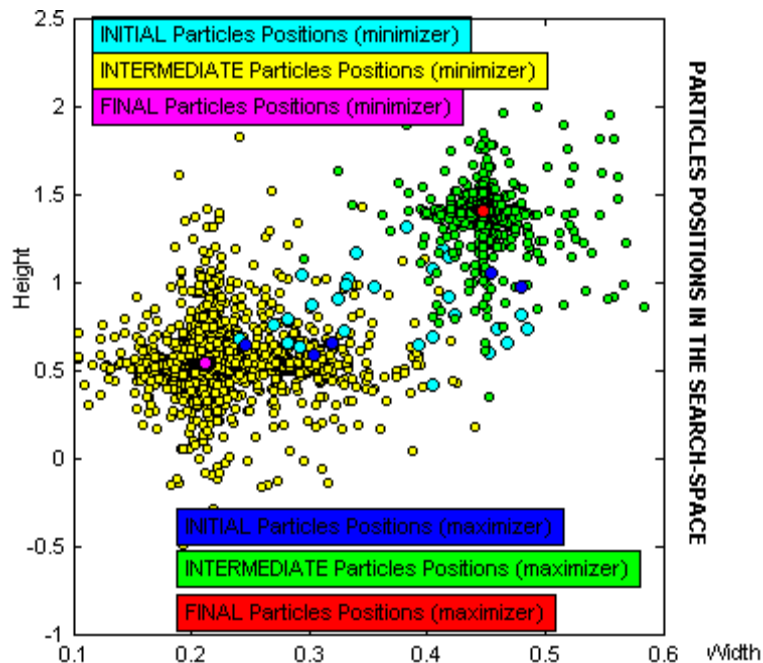

**Fig. 1** Evolution of the particles' positions over the search-space for the first optimizer and CASE A. The minimizer finds the solution, while the maximizer is used for the termination conditions only.

It must be noted that the penalization needed to be increased to a constant value equal to $k = 10^8$ for the second optimizer in **Table 3** in order to sort out some constraint violation.

## 7   CONCLUDING REMARKS AND FUTURE WORK

A procedure for the section design of FRP-reinforced concrete beams following the recommendations in ACI.440.1 R-06 [1] was presented along section 2, and the optimization of the flexural section design was successfully carried out making use of the particle swarm optimization paradigm. Two optimizers were proposed, which differ from one another in two important aspects:

1.   The first optimizer uses the preserving feasibility technique to deal with the constraints—which guarantees the feasibility of the solution found—, while the second one uses the penalization method.

2.   The first optimizer computes the reinforcement in a deterministic fashion at each time-step, while the second one considers the reinforcement as yet another object variable to be optimized.

The implications of these two aspects are of utmost importance. Although the preserving feasibility technique guarantees the feasibility of the solution, it requires that all the particles are initialized within feasible space. In addition, it neglects the information contained in the infeasible space to direct the search, which can be an important issue for problems where the feasible space is very small in with respect to the whole search-space. It is fair to note that the results obtained by the first optimizer in **Table 3** required notably greater resources due to the deterministic calculation of the reinforcement, although a personal computer and a few minutes are all that is required. Nonetheless, both optimizers proved themselves able to find near-optimal solutions, as it was verified manually.

The main drawback of the penalization method is that it requires the problem-dependent tuning of the penalization coefficients, where too high penalizations lead to sub-optimal solutions and too low ones lead to infeasible solutions. The research on the design of adaptive penalizations is currently ongoing.

Since the smaller diameters of the bars present higher tensile strengths, the deterministic calculation of the reinforcement is implemented as an iterative process that starts with the smallest diameter, which is iteratively increased until all the constraints are satisfied. Since the maximum feasible width was set to a high value (100 cm), the design returned for CASE B by the first optimizer in **Table 3** exhibits a wide width and numerous bars of small diameters.

As a first step towards a more sophisticated model, the design here is limited to section flexural design, and to uniform diameter selection. Note, however, that extending the second optimizer to the use of different diameters is relatively straightforward: the addition of a few dimensions is all that is needed. Another advantage of considering the reinforcement as object variables rather than calculating it in a deterministic fashion is that the addition, removal or modification of constraints—for instance to include the shear design—is notably easier. Thus, the second optimizer appears to be a further step towards an optimizer able to deal with more complex optimal designs.

## ACKNOWLEDGMENTS

The authors would like to gratefully acknowledge the support provided by the Spanish Government (Ministerio de Educación y Ciencia), Project Ref. BIA-2004-05253.